\documentclass[10pt,twocolumn,letterpaper]{article}

\usepackage[pagenumbers]{cvpr} 

\usepackage{graphicx}
\usepackage{amsmath}
\usepackage{amssymb}
\usepackage{booktabs}
\usepackage{array,multirow}
\usepackage{algorithm}
\usepackage{algorithmicx}
\usepackage[noend]{algpseudocode}
\usepackage{cuted}
\usepackage{capt-of}
\usepackage[marginal]{footmisc}
\usepackage[accsupp]{axessibility}

\usepackage[pagebackref,breaklinks,colorlinks]{hyperref}

\usepackage[capitalize]{cleveref}
\crefname{section}{Sec.}{Secs.}
\Crefname{section}{Section}{Sections}
\Crefname{table}{Table}{Tables}
\crefname{table}{Tab.}{Tabs.}

\def\cvprPaperID{402} 
\def\confName{CVPR}
\def\confYear{2022}

\begin{document}


\title{HumanNeRF: Efficiently Generated Human Radiance Field from Sparse Inputs}

\author{Fuqiang Zhao$^{1}$
\and
Wei Yang$^{2}$
\and
Jiakai Zhang$^1$
\and
Pei Lin$^1$
\and
Yingliang Zhang$^{3}$
\and
Jingyi Yu$^1$ \qquad \qquad Lan Xu$^{1,4}$\\
$^{1}$ ShanghaiTech University \qquad $^{2}$ Huazhong University of Science and Technology \qquad $^{3}$ DGene \\
\qquad $^{4}$ Shanghai Engineering Research Center of Intelligent Vision and Imaging
}

\maketitle
\begin{strip}\centering
\includegraphics[width=1.0 \textwidth]{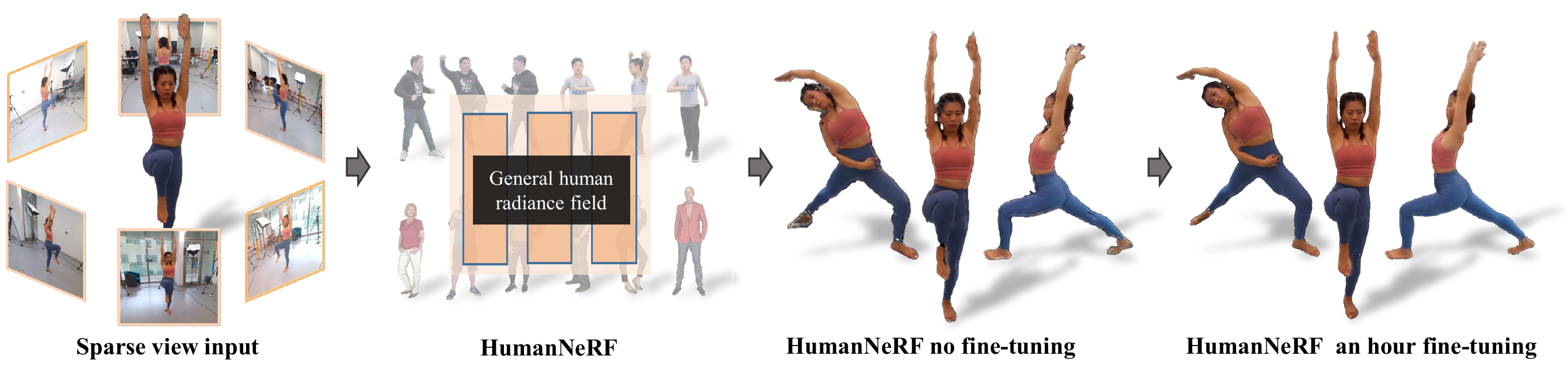}
\captionof{figure}{Our proposed HumanNeRF utilizes on-the-fly efficient general dynamic radiance field generation and neural blending, enabling high-quality free-viewpoint video synthesis for dynamic humans. Our approach only takes sparse images as input and uses a pre-trained network on large human datasets. Then we can effectively synthesize a photo-realistic image from a novel viewpoint. While these results contain artifacts, we fine-tune 300 frames for a specific performer using only an hour and generate improved results.}
\label{fig:teaser}
\end{strip}

\begin{abstract}
    Recent neural human representations can produce high-quality multi-view rendering but require using dense multi-view inputs and costly training. They are hence largely limited to static models as training each frame is infeasible. We present HumanNeRF - a neural representation with efficient generalization ability - for high-fidelity free-view synthesis of dynamic humans. Analogous to how IBRNet assists NeRF by avoiding per-scene training, HumanNeRF employs an aggregated pixel-alignment feature across multi-view inputs along with a pose embedded non-rigid deformation field for tackling dynamic motions. The raw HumanNeRF can already produce reasonable rendering on sparse video inputs of unseen subjects and camera settings. To further improve the rendering quality, we augment our solution with in-hour scene-specific fine-tuning, and an appearance blending module for combining the benefits of both neural volumetric rendering and neural texture blending. Extensive experiments on various multi-view dynamic human datasets demonstrate effectiveness of our approach in synthesizing photo-realistic free-view humans under challenging motions and with very sparse camera view inputs.
\end{abstract}

\section{Introduction}
\label{sec:intro}
View synthesis of human activities enables numerous applications in visual effects and telepresence, with unique and immersive viewing experiences.
However, a convenient and high-quality solution from the light-weight capture setup remains a cutting-edge yet bottleneck technique.

Early solutions require a dome-based multi-view setup for accurate reconstruction~\cite{collet2015high,dou-siggraph2016} and image-based rendering in novel views~\cite{carranza2003free,zitnick2004high}.
Volumetric approaches~\cite{xu2019unstructuredfusion,su2021robustfusion} enable light-weight reconstruction, but they still heavily rely on the depth sensors and are restricted by the limited mesh resolution. 
Recent neural rendering techniques have achieved significant progress~\cite{NR_survey,NeuralVolumes,mildenhall2020nerf,habermann2021real}.
Remarkably, NeRF~\cite{mildenhall2020nerf} and its dynamic extensions~\cite{pumarola2020d,peng2021neural,Tretschk_2021_ICCV,wang2021ibutter,liu2021neuralActor,STNeRF_SIGGRAPH2021} enable photo-realistic novel view synthesis for dynamic scenes without heavy reliance on the reconstruction accuracy.
However, these solutions still require expensive dense capture views or suffer from tedious time-consuming per-scene training,  which highly limits the practicality. 
Only recently, some approaches~\cite{yu2021pixelnerf,wang2021ibrnet,chen2021mvsnerf} enhance NeRF~\cite{mildenhall2020nerf} with image-conditioned features to break the per-scene training constraint for efficient radiance field generation of static scenes.
But few researchers explore such generalizable NeRF representation under the complex dynamic human settings.
The recent work~\cite{NeuralHumanFVV2021CVPR} further enables generalizable human rendering from 6 RGB streams by combining texture blending with implicit geometry inference~\cite{saito2019pifu,saito2020pifuhd} only in novel views.
However, it suffers from severe artifacts near the occluded regions due to the lack of global inherent geometry and texture modeling.
%

In this paper, we present \emph{HumanNeRF} -- a practical and high-quality neural free-view synthesis approach for general dynamic humans using only sparse RGB streams. 
As illustrated in Fig.\ref{fig:teaser}, our approach enables photo-realistic human rendering by efficiently optimizing a more generalizable radiance field on-the-fly for unseen performers in an hour, favorably transcending previous long-term per-scene training approaches.

%
Our key idea is to marry the dynamic NeRF representation with neural image-based blending in a light-weight and two-stage framework. 
We extend the concept of general radiance field into the dynamic and temporal setting to break the per-scene constraint for efficient rendering.
We also explore an effective implicit blending strategy to boost the texture result of volumetric rendering with the level of detail present in the sparse input images.
Specifically, we first adopt an implicit scheme to aggregate image-conditioned features from our sparse input, which enables generalizable inference of motion and appearance in the dynamic NeRF framework.
Then, we introduce a pose-embedded hybrid deformation scheme to enhance the generalization ability for unseen identities under various motions and garments. 
It combines explicit model-based warping with implicit subtle displacement modeling, so as to learn a reliable radiance field in an inherent canonical space.
Note that our scheme also supports efficient per-performer fine-tuning with temporally sparse sampling, which significantly improves the rendering quality even on unseen poses.
However, we observe that existing dynamic NeRF-based volumetric rendering still fails to generate high-frequency texture details, especially for challenging unseen identities and poses. To this end, we combine the image-based rendering with NeRF-based volume rendering into a novel neural blending scheme through implicit and occlusion-aware blending weight learning. 
It enables accurate appearance rendering in the target view with the level of texture detail in the adjacent input images. 
To summarize, our main contributions include:
\begin{itemize}
	\item We present a high-quality performance rendering approach via efficient radiance field generation for arbitrary performers from sparse RGB streams, achieving significant superiority to existing the state of the art.
	
	\item We extend the generalizable NeRF into the new realm of dynamic and light-weight setting through implicit feature aggregation and hybrid deformation.

	\item We propose a novel implicit blending scheme to preserve the texture detail from the input images, providing photo-realistic appearance rendering.
	
\end{itemize}

\section{Related work}
\label{sec:Relatedwork}
\textbf{Human Performance Capture.}
Markerless human performance capture techniques have been widely adopted to achieve human free-viewpoint video or reconstruct the geometry. 
Some recent work only relies on the light-weight and single view setup \cite{xu2018monoperfcap,habermann2019livecap,xu2020eventcap,chen2021sportscap}, but these methods require the pre-scanned template or naked human model and it is difficult for them to achieve photo-realistic view synthesis.
The high-end approaches \cite{stoll2011fast,liu2013markerless,liu2021neuralActor,habermann2021real}  are able to produce high-quality surface motion and appearance reconstruction, but they require dense cameras and a controlled imaging environment which is not easily accessible.
Other monocular RGB-D based methods \cite{newcombe2015dynamicfusion,guo2017real,yu2018doublefusion,xu2019flyfusion,su2020robustfusion, jiang2022neuralfusion} adopt the traditional modeling and rendering pipeline to synthesize novel views of humans. However, these methods still suffer from the inherent self-occlusion constraint and cannot capture the motions in occluded regions.
The light-weight multi-view solutions \cite{dou-siggraph2016,dou2017motion2fusion,xu2019unstructuredfusion} which is most similar to our method serve as a good compromise between over-demanding hardware setup and high-fidelity reconstruction but still rely on 3 to 8 RGBD streams as input.
%
\begin{figure*}[t]
	\begin{center}
		\includegraphics[width=0.95\linewidth]{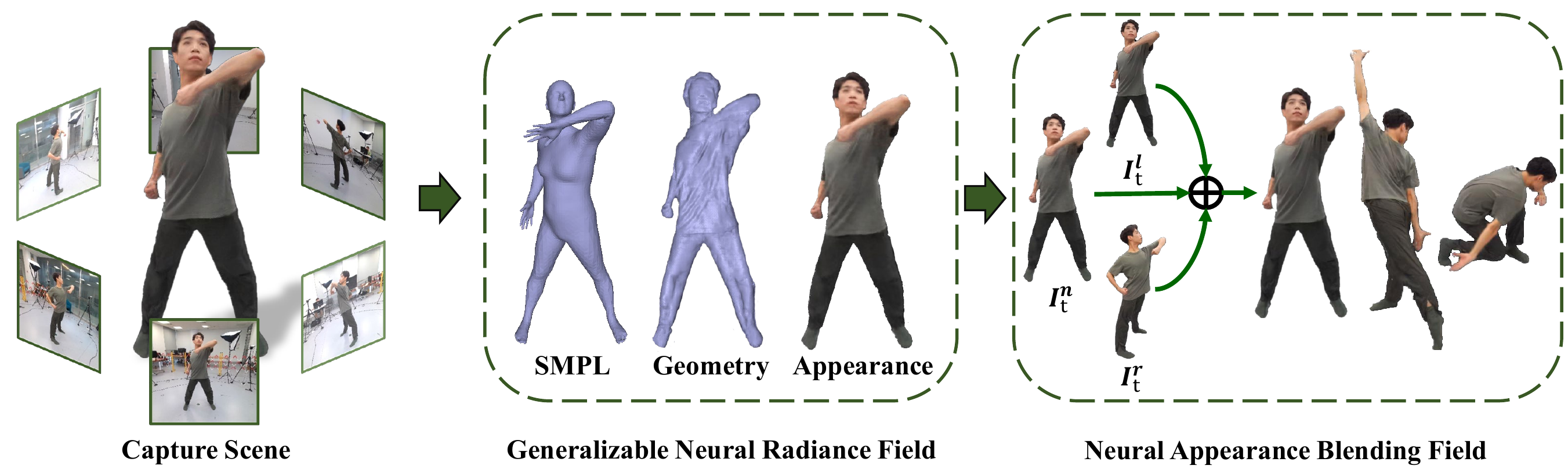}
	\end{center}

	\caption{The overview of our HumanNeRF method. Assuming the video input from six RGB cameras surrounding the performer, our approach consists of a generalizable neural radiance field (Sec.~\ref{sec:generative}), an optional fast per-scene fine-tuning scheme and a novel neural appearance blending field (Sec.~\ref{sec:refinement}).}
	\label{fig:overview}

\end{figure*}

\textbf{Neural Rendering.}
Recently, a lot of work has shown significant progress on 3D scene modeling and photo-realistic novel view synthesis via differentiable neural rendering manner based on various data representations, such as point clouds \cite{aliev2020neural,wu2020multi}, voxels \cite{yan2016perspective,sitzmann2019deepvoxels,NeuralVolumes},or texture meshes \cite{thies2019deferred,liu2019neural}.
More recent implicit manner based work \cite{sitzmann2019scene,mildenhall2020nerf,liu2020neural,peng2021animatable, tiwari2021neural,su2021nerf,xu2021h} achieves impressive results for novel view synthesis for a specific scene.
However, dedicated per-scene training is required in these methods when applying the representation to a new scene.  
Some methods~\cite{kwon2021neural,wang2021ibrnet,chen2021mvsnerf,NeuralHumanFVV2021CVPR, saito2019pifu, riegler2021stable} utilize pixel-aligned features from source images to enable generalizable human modeling without per-scene training constraint.
However, the method~\cite{saito2019pifu} generates blur texture results due to the reliance on implicit texture representation, while the method~\cite{NeuralHumanFVV2021CVPR} suffers from geometric discontinuity due to the lack of temporal information. 
Recently, Kwon~\textit{et al.}~\cite{kwon2021neural} utilized temporally aggregated features to compensate the sparse input views, achieving generalizable human radiance field generation.
However, they still suffer from blur artifacts when generalizing unseen identities with complex motions due to self-occlusion.
In contrast, we utilize generalizable human NeRF with occlusion-aware pixel-aligned features and adopt implicit blending, achieving high-quality novel view synthesis with the level of texture detail present in the input images.

\textbf{Image based rendering.}
Previous work of IBR \cite{debevec1996modeling,gortler1996lumigraph,levoy1996light} aims at synthesizing novel view from a set of source images through blending weights of reference pixels without recovering detailed 3D geometry. Blending weights are calculated based on ray space approximation \cite{levoy1996light} approximate proxy geometry \cite{debevec1996modeling,buehler2001unstructured, heigl1999plenoptic}.
Though their rendering results are impressive, the range of renderable viewpoints is limited.
In recent work\cite{zitnick2004high,chaurasia2013depth,penner2017soft}, researchers have proposed improved methods by inferring depth maps from input images as proxy geometries. 
For example, some work \cite{hedman2018deep,suo2021neuralhumanfvv} utilize two stages of multi-view stereo. First, they generate a grid surface that depends on the view and then there is a CNN to calculate the blending weights.
While these methods can handle sparser views than other approaches and achieve promising results in some cases, they are sensitive to the quality of reconstructed proxy geometries \cite{jancosek2011multi,saito2019pifu}.
Comparably, our method embraces image blending into implicit representations pipeline under the light-weight multi-RGB, which enables photo-realistic appearance and geometry reconstruction in novel views.

\begin{figure*}[t]
	\begin{center}
		\includegraphics[width=1.0\linewidth]{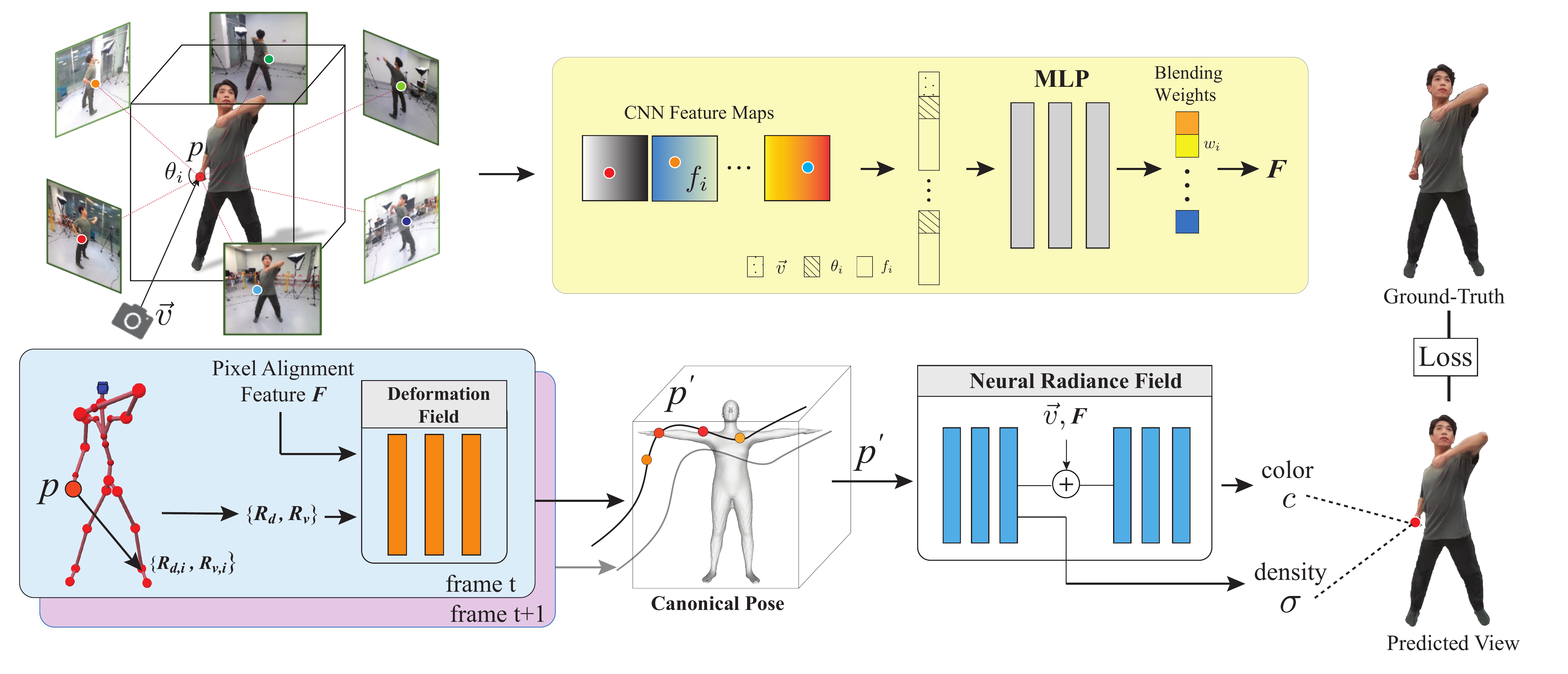}
	\end{center}
	\caption{Illustration of our generalizable neural radiance field module. For feature extraction, we concatenate $\{f_i\}^{\mathcal{K}}_{i=1}$ with the viewing directions $\vec{v}$ and the angle $\{\theta_i\}^{\mathcal{K}}_{i=1}$ of viewing direction relative to each the query ray of the source view. We use $R_d$ and  $R_v$ to model the distances and directions between sample point $p$ and the 24 joints of the SMPL skeleton. We use the neural radiance field to regress volume density and RGB radiance at location $p'$ in canonical space and its corresponding blending features $F$.}
	\label{fig:pipeline}
\end{figure*}

\section{The HumanNeRF Approach}
We first introduce the problem formulation and overall scheme of our HumanNeRF method. Given $K$ synchronized videos of a performer captured at different viewpoints (360\textdegree around preferably) with $T$ frames, $\mathbf{\mathcal{I}} = \{I^{k,t}\}$, in each video, our method aims to synthesize free-viewpoint videos of the performer and also generalize the motion to an arbitrary person with high fidelity. Fig.~\ref{fig:overview} illustrates the high-level components of our system. The core step of our approach is the efficient generalized Neural Radiance Field for dynamic humans, which adapts the NeRF~\cite{mildenhall2020nerf} for dynamic human representation.

We leverage the parametric human body model SMPL \cite{loper2015smpl} for estimating a basis model, and use an MLP network to learn subtle displacements of the human body. The output is then deformed into a canonical pose for NeRF optimization and rendering (Sec.~\ref{sec:generative}). The efficient generalization ability comes from our aggregated pixel alignment features $\mathbf{\mathcal{F}} = \{F_t\}$ from multi-view input images by projecting the 3D sample point into images and blending individual image features. 
Although our generated NeRF outputs human geometry with good quality, synthesized textures may contain artifacts and lack high-frequency details. Thus, we use a novel neural appearance blending scheme to refine texture details by aggregating colors from neighboring views. The final synthesized results exhibit a photo-realistic appearance with fine details (Sec.~\ref{sec:refinement}).

\subsection{Generalizable Dynamic Neural Radiance Field}
\label{sec:generative}

We retain NeRF's ability for novel view synthesis and geometry details rendering. However, NeRF assumes stationary subjects and performs per-scene optimization, which makes it not directly applicable to our problem. 
We make two main changes to NeRF to handle human dynamics and gain generalization ability. Specifically, we first warp the camera ray to account for human motion before sampling for NeRF and combine the viewing direction input with aggregated pixel alignment features for gaining generalizability.

\textbf{Aggregated Pixel Alignment Feature.}
We propose an aggregated pixel alignment feature for NeRF generalization. Specifically, we use a U-Net network $U$ to extract image feature maps representing local image appearance. Given an input image $I^{k} \in \mathbb{R}^{H\times W\times  4}$ with mask as the last channel, the output of $U$ is a 2D feature map $f^{k} \in \mathbb{R}^{H\times  W\times  C}$, i.e.,

\begin{equation}
	f^{k} = U(I^{k}) 
\end{equation}
For each spatial point $p \in \mathbb{R}^3$ fed to NeRF, we first project it into view $k$ at $q^k \in \mathbb{R}^2$ and fetch the corresponding feature vector $f^{k}_q$. The aggregated pixel alignment feature of $p$ then is a weighted summation of image features as $F_p = \sum^{K}_{k=1} w^{k} f_q^{k}$ for $k$ in $1...K$, and we use an MLP network to estimate the blending weights $w^{k}_q$ as: 
\begin{equation}
	w^{k}_q = \mathbf{MLP}_{\mathcal{B}} (\vec{v}, \theta^k_q, f^{k}_q) 
\end{equation}
\noindent where $\vec{v}$ is the view direction of $p$ in camera view and $ \theta^k_q$ is the angle of the viewing direction w.r.t. the sample ray from $p$ to $q^k$.

\textbf{Pose Embedded Non-rigid Human Deformation.}
To accommodate the human dynamics, we warp the human body from the current time frame to a common canonical pose so that NeRF receives static sampling queries, similar to \cite{pumarola2020d,park2020deformable,tretschk2021non,liu2021neuralActor}. In practice, we find an MLP module tends to learn subtle displacements other than handle large deformations. To address this issue, we fit the SMPL model to a human body in the current time frame and deform the model to a common canonical pose using inverse-skinning transformation $\mathcal{S}$ \cite{huang2020arch,liu2021neuralActor}. The resulting model usually exhibits inconsistencies with image observations. We further apply a pose-dependent non-rigid deformation field $\mathbf{MLP}_d$ to learn the subtle displacement. Our pose embedded non-rigid deformation field can be formulated as:
\begin{equation}
    p'= \mathcal{S}(p, \mathcal{M}, w^s) + \mathbf{MLP}_d (R_d, R_v, F_p)
\end{equation}
\noindent where $\mathcal{M}$ is the estimated motion, $w^s$ is the corresponding skinning weight of sample point $p$. We use $R_d \in \mathbb{R}^{24}$ and $R_v \in \mathbb{R}^{72}$ to model the distances and directions between $p$ and the 24 joints of SMPL skeleton. $F_p$ is the aggregated pixel alignment feature. 

Finally, we have our generalizable dynamic neural radiance field $\Phi$, which takes the transformed 3D location $p'$, view direction $\vec{v}$ and $F_p$ as input and predicts the volume density $\sigma$ and color $c$ at point $p$ before deformation as:

\begin{equation}
    (\mathbf{c}, \sigma) = \Phi(p', \vec{v}, F_p)
\end{equation}

Fig.~\ref{fig:pipeline} shows the overview of our generalizable dynamic neural radiance field. 

\textbf{Dynamic Human Volume Rendering.}
We utilize the physically based volume rendering \cite{kajiya1984ray} technique to synthesize a new view image similar to the original NeRF. The only difference here is the query ray is bent by the deformation field before sending to NeRF. In particular, we compute a pixel’s color at frame $t$ by marching a corresponding ray and accumulating radiance at sampled points between near and far bounds, i.e.,

\begin{equation}
    C_\mathbf{r} = \sum^{N}_{i=1}T(p_i) \big [(1 - e^{-\sigma_{p_i}\delta_{p_i}}) c_{p_i} \big]
\end{equation}
where $T(p_i) =  e^{-\sum^{i-1}_{k=1}\sigma_{p_i}\delta_{p_i}}$ and $\delta(p_i) = p_{i+1} - p_{i}$ is the distance between adjacent samples and $N$ is the number of the sampled points on the ray.

\textbf{Fast Per-subject Fine-tuning.}
Due to the limited training data along with diversity among different identities and scenes, artifacts and imperfections can still be observed for an unseen person when transferring motion. 
To address this problem, we adopt a fast fine-tuning solution as compensation to our original framework which treats the network optimized on the performer as an initialization state. Specifically, we first train our network on various subjects/performers and freeze the feature blending network $\mathbf{MLP}_{\mathcal{B}}$. And then when given an unseen subject, we optimize the network parameters of our deformation field $\mathbf{MLP}_d$ and the generalizable neural radiance field $\Phi$.

\subsection{Neural Appearance Blending} \label{sec:refinement} 
We observe that textures produced by NeRF rendering in the above section contain artifacts and lack high-frequency details sometimes due to the sparsity of input views. Inspired by image based rendering methods, we further propose a novel neural blending scheme for appearance refinement. Most of the texture information in a target view can be recovered by its only two adjacent input views in our multi-view setting. Considering a certain time frame, we first render a depth map $D^\mathbf{v}$ at the target view $\mathbf{v}$ from our generalizable neural radiance field at inference time. And then we back-project each point $q$ in $D^\mathbf{v}$ with color $C^\mathbf{v}_q$ into neighboring two views $\mathbf{v}_1$ and $\mathbf{v}_2$ and fetch the colors $C^{\mathbf{v}_1}_q$ and $C^{\mathbf{v}_2}_q$, along with the corresponding visibility $O^{\mathbf{v}_1}_q$ and $O^{\mathbf{v}_2}_q$ which is determined by depth difference.
At training time, the depth maps are rendered from the synthetic human model dataset, such as Twindom \cite{twindom}. 
We further extract $q$'s corresponding image feature $f^{\mathbf{v}_1}_q,f^{\mathbf{v}_2}_q$, and then feed feature and visibility information into our neural appearance blending network $\mathbf{MLP}_\mathcal{A}$,

\begin{equation}
    \begin{aligned}
        W_q = \mathbf{MLP}_\mathcal{A}(f^{\mathbf{v}_1}_q, O^{\mathbf{v}_1}_q, f^{\mathbf{v}_2}_q, O^{\mathbf{v}_2}_q)
    \end{aligned}
\end{equation}
\noindent where $W_q \in \mathbb{R}^3$ is the appearance blending weight. The final color for $q$ in $\mathbf{v}$ then is:
\begin{equation}
    \begin{aligned}
        C^{\mathbf{v}*}_q = W_q \cdot C_q, \ \ C_q = [C^{\mathbf{v}_1}_q, C^{\mathbf{v}}_q, C^{\mathbf{v}_2}_q]
    \end{aligned}
\end{equation}
\noindent where $\cdot$ denotes the dot product.

\begin{figure}[t]
	\begin{center}
		\includegraphics[width=1.0\linewidth]{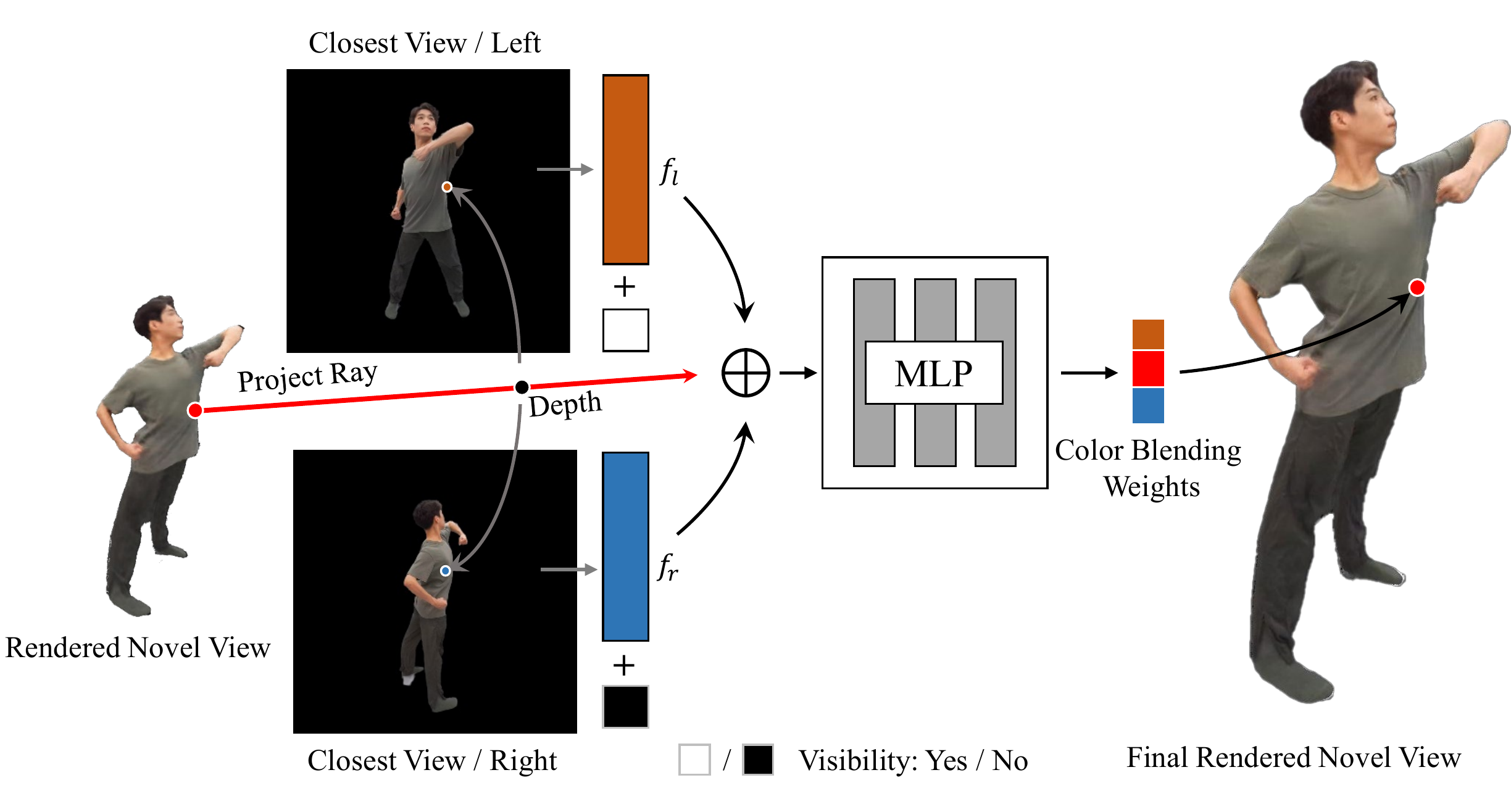}
	\end{center}
	\caption{Illustration of our neural appearance refinement scheme. Our appearance blending network takes two adjacent image features $f_r,f_l$ and corresponding occlusion information as input and then output three-dimensional weights to blend our rendering result with fine-detailed appearance information of the adjacent input views.}
	\label{fig:Sampling}
\end{figure}

\begin{figure*}[t]
\begin{center}
    \includegraphics[width=0.95\linewidth]{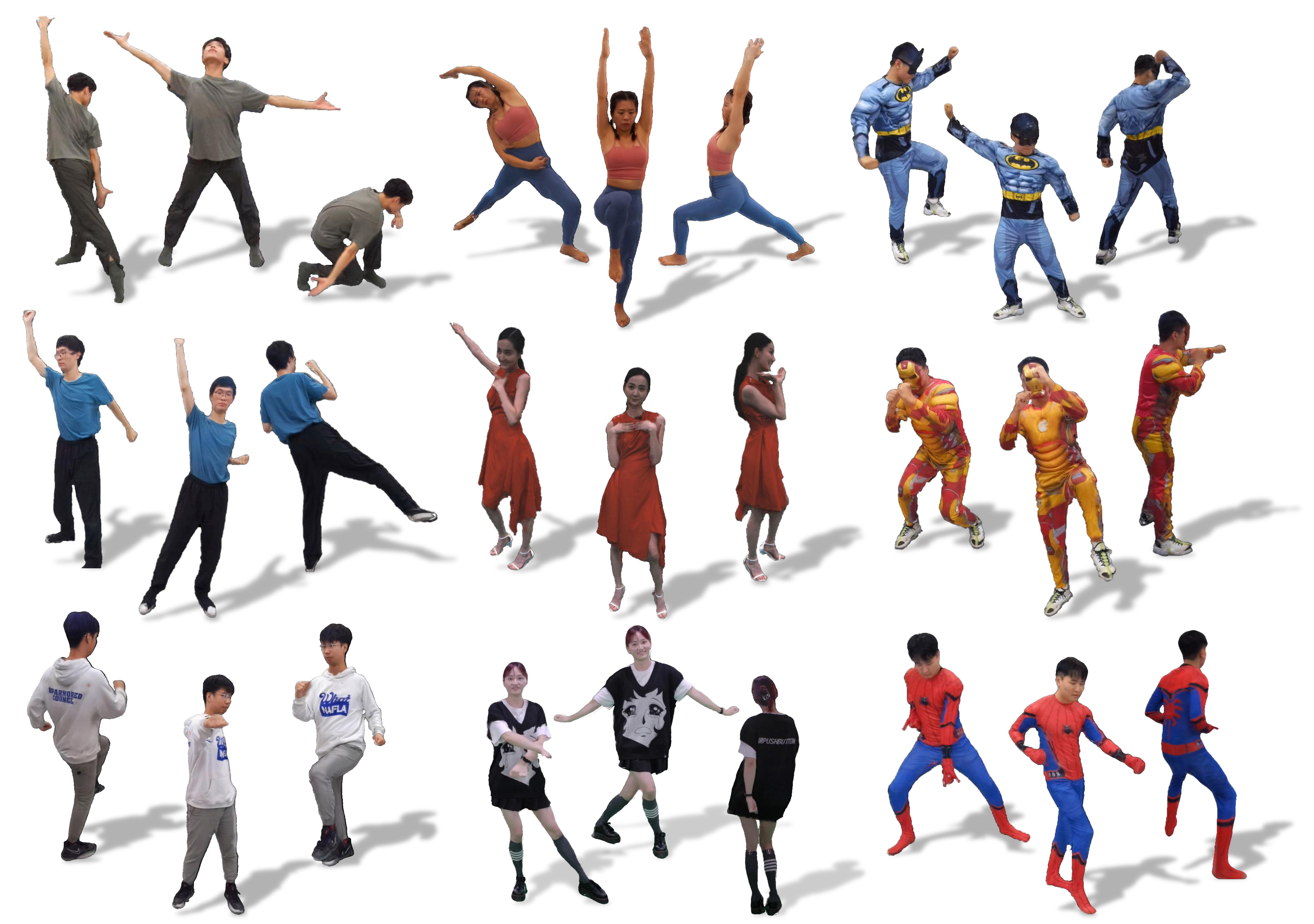}
\end{center}

\caption{The appearance results of our HumanNeRF method on several sequences, including “dance1”, “yoga”, “batman”, "swing", "dance2", "ironman", “sport”, “dance3” and “spiderman” from the upper left to lower right.}
\label{fig:results}

\end{figure*}

\subsection{Implementation Details}
Here we describe the implementation details including the training scheme of our approach. Our generalizable NeRF module (including the feature extraction network $U$, feature blending network $\mathbf{MLP}_\mathcal{B}$, deformation network $\mathbf{MLP}_d$ and the adapted NeRF $\Phi$) and the appearance blending network $\mathbf{MLP}_\mathcal{A}$ are independent and we train them separately.

To optimize our network, we use a color loss $\mathcal{L}_{c}$  which measures the difference between the rendered color $C_\mathbf{r}$ and the ground truth color $\hat{C}_\mathbf{r}$ of camera ray $\mathbf{r}$:
\begin{equation}
\mathcal{L}_{c} = \sum_{\mathbf{r} \in \mathcal{R}}(||C_\mathbf{r} - \hat{C}_\mathbf{r}||^2_2)
\end{equation}
\noindent and a silhouette loss $\mathcal{L}_{m}$ which is formulated as:
\begin{equation}
\mathcal{L}_{m} = \sum_{\mathbf{r} \in \mathcal{R}} \mathbf{BCE}(M_\mathbf{r} - \alpha_\mathbf{r})
\end{equation}
\noindent where $\alpha(\mathbf{r}) = \sum^{N}_{i=1}T(p_i)(1 - e^{-\sigma_{p_i}\delta_{p_i}})$ is the rendered mask for $\mathbf{r}$. The total loss is combination of $\mathcal{L}_{c}$ and $\mathcal{L}_{m}$:
\begin{equation}
\mathcal{L} = \mathcal{L}_{c} + \lambda \mathcal{L}_{m}
\end{equation}
where $\lambda$ is the weight to balance the two losses. Specifically, we set $\lambda$=0.1 in our implementation. And we only use $\mathcal{L}_c$ for the neural appearance blending model.

\textbf{Training Details.}
We train our models using Adam optimizer with a learning rate that decays from $1e^{-4}$ to $1e^{-5}$ during training. Besides, we sample $4,096$ camera rays for each mini-batch and sample 32 and 64 points from near to far following the hierarchical sampling strategy.
 We optimize all our networks on a PC with a single Nvidia GeForce RTX3090 GPU. The training time of our generalizable NeRF module is about 2 days. Depending on the number of video frames, the fine-tuning time ranges from 30 to 90 minutes, for input images with $1,080 \times 1,080$ resolution. Besides, we train our neural appearance refinement model for about 1 to 2 days. 

 \textbf{Datasets.} We train our generalizable NeRF on 1820 static scans from the Twindom \cite{twindom} dataset, which consists of 120 camera views. And we collect 6 view videos for 26 subjects with challenge motion, such as dancing and yoga. We also augment the data by rigging the pre-scanned 3D model and simulating challenging poses with 120 views to boost the generation ability of our networks. For the neural appearance blending module, we only train it on Twindom\cite{twindom} dataset.
 
\section{Experimental Results}

In this section, we evaluate our HumanNeRF method on a variety of challenging scenarios. As demonstrated in Fig.~\ref{fig:results}, our approach generates high-quality appearance results and handles humans with rich textures, challenging poses and etc.

\subsection{Comparison}
We first compare our HumanNeRF method with per-scene optimization approaches including Neural Body \cite{peng2021neural}, Neural Volumes \cite{NeuralVolumes} and ST-NeRF \cite{STNeRF_SIGGRAPH2021} both qualitatively and quantitatively. Furthermore, we also compare our method with generalizable methods, i.e., IBRNet \cite{wang2021ibrnet} and NeuralHumanFVV \cite{NeuralHumanFVV2021CVPR}, in our sparse view input setting.

\begin{figure*}[t]
 	\begin{center}
 		\includegraphics[width=0.95\linewidth]{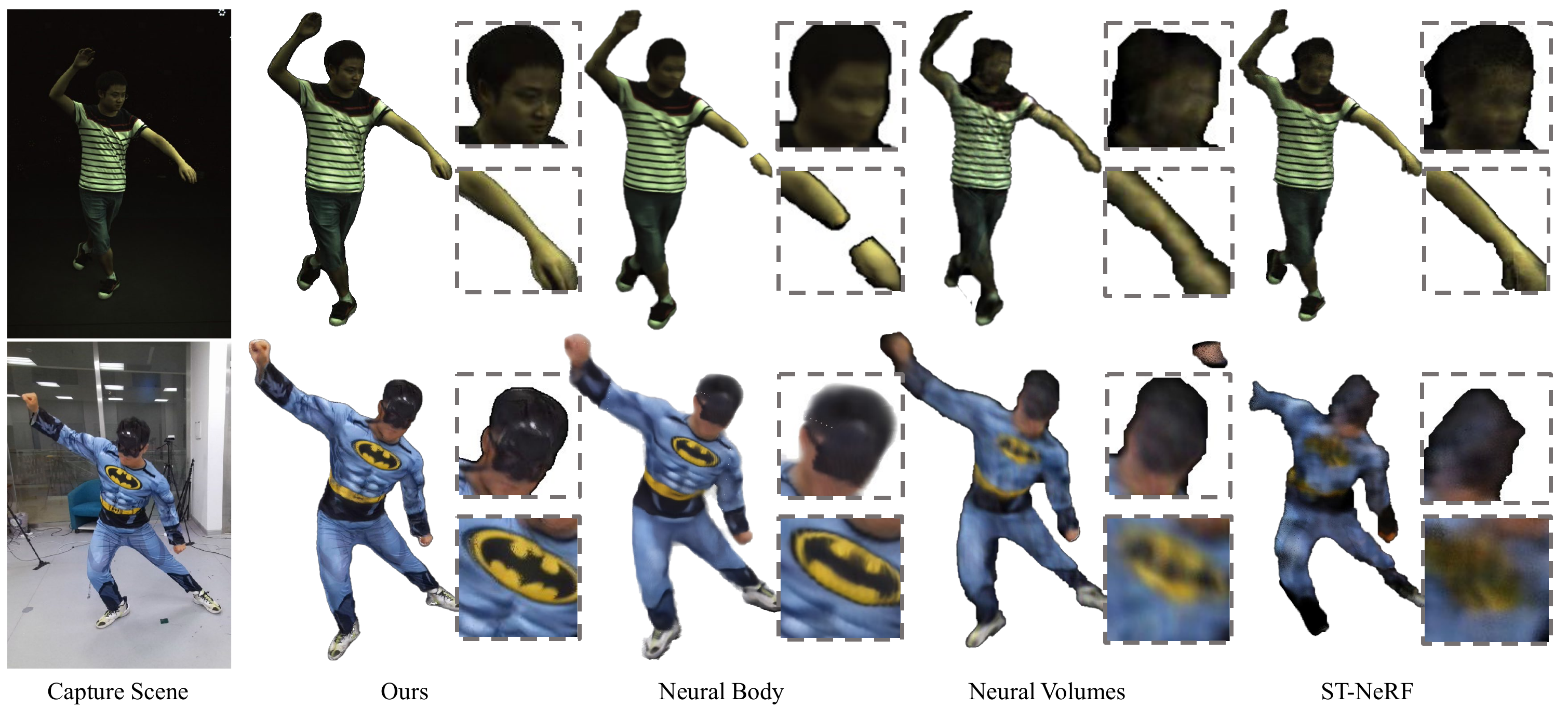}
 	\end{center}
 	\vspace{-0.4cm}
 	\caption{Qualitative comparison against per-scene training methods. We compare our method with Neural Body, Neural volumes, and ST-NeRF on “Batman” from our multi-view datasets and “Taichi” from ZJU-MoCap datasets. Our approach generalizes the most photo-realistic and finer detail.}
 	\label{fig:comparison perscene}
 \end{figure*}

\begin{figure}[t]
 	\begin{center}
 		\includegraphics[width=0.95\linewidth]{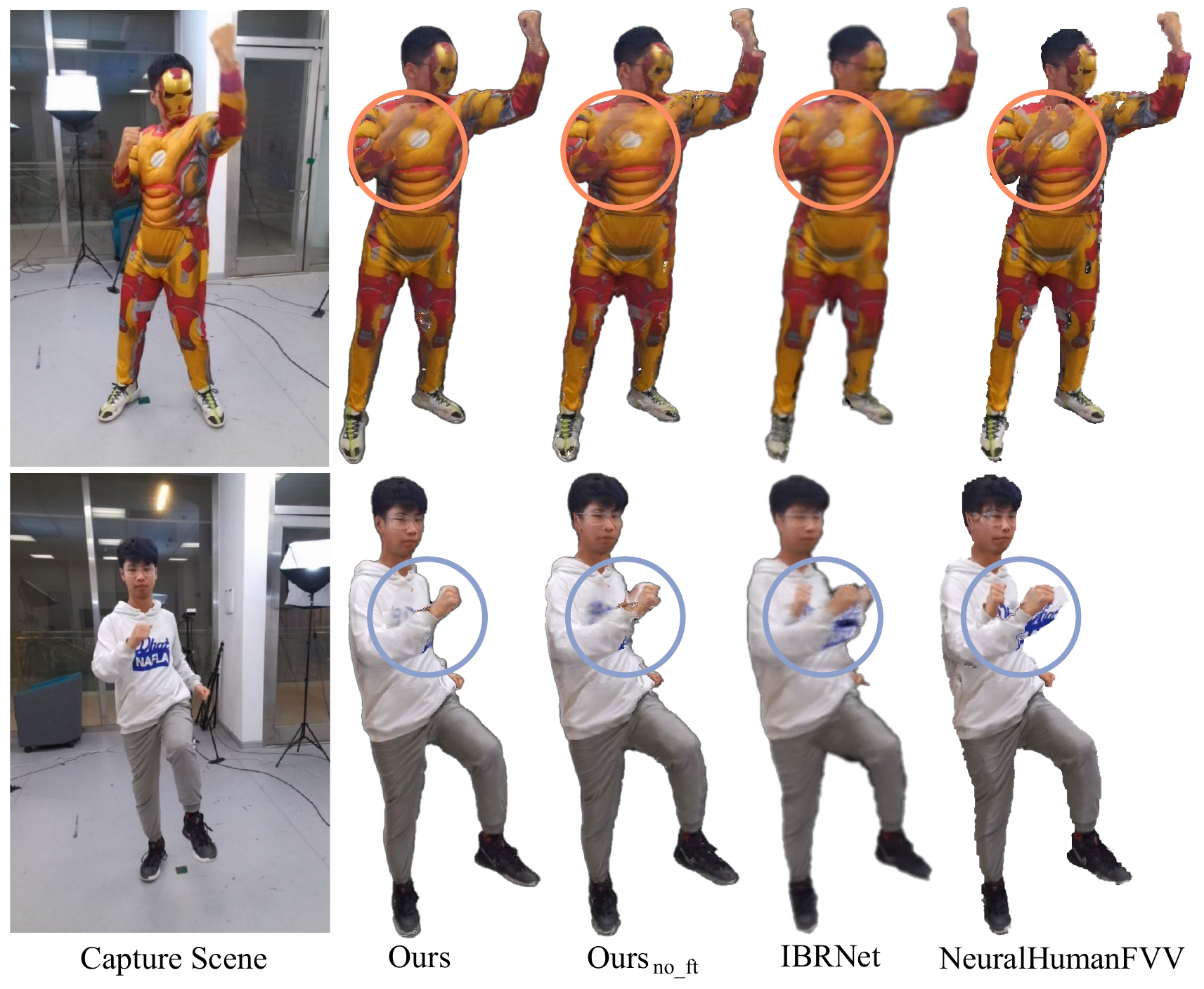}
 	\end{center}
 	\caption{Qualitative comparison against generalizable methods. We compare our method with IBRNet, NeuralHumanFVV. Note that our approach generates better appearance results and well addresses the self-occlusion problem.}
 	\label{fig:comparison multiscene}

 \end{figure}

As shown in Fig.~\ref{fig:comparison perscene}, 
compared with the per-scene optimization methods, our HumanNeRF achieves better results in just a short fine-tuning time. Results from our approach exhibit much better textures and the geometries are complete and accurate both for the ``Taichi'' from public ZJU-MoCap \cite{peng2021neural} and the ``Batman'' data collected by ourselves. When compared with generalizable methods, our method outperforms others and well addresses self-occlusions as shown in Fig.~\ref{fig:comparison multiscene}. 

As for quantitative comparison, we show the PSNR, SSIM, LPIPS and MAE metrics of our approach and other methods on real testing data in Tab.~\ref{table:quantitative comparison}. Specifically, we set reference camera images as ground truth, and calculate the metrics for synthesized images from methods for comparison.

 \begin{table}[t]
 	\centering
 	\begin{tabular}{l|c|c|c|c}
 		Method &  PSNR$\uparrow$ & SSIM$\uparrow$&LPIPS$\downarrow$ &MAE$\downarrow$ \\ \hline
 		ST-NeRF & 17.34 &  0.8547 & 0.1493& 11.38\\
 		NeuralVolumes &  27.32&   0.9408& 0.0999& 2.905 \\
 		NeuralBody & 28.21 & 0.9544  & 0.0762& 2.294\\ \hline
 		IBRNet & 30.73 & 0.9817 & 0.0348 & 1.154\\ 
 		NeuralHumanFVV & 27.86 & 0.9785  & 0.0440 & 1.237\\ \hline
 		Ours$_{wo\_bo}$ & 25.80 &  0.9456 & 0.0825& 3.354\\  
 		Ours$_{wo\_ft}$ & 29.51 &  0.9741 & 0.0461& 1.521\\
 		Ours$_{wo\_rf}$ & 29.69 &  0.9620 & 0.0703& 2.016\\
 		Ours & \textbf{33.01} &  \textbf{0.9842} & \textbf{0.0334} & \textbf{0.9307}\\    \hline
 	\end{tabular}
 	\rule{0pt}{0.05pt}
 	\caption{\textbf{Quantitative comparison against several methods in terms of rendering accuracy.} Compared with NeRF, ST-NeRF, Neural Volumes, NeuralBody, IBRNet and NeuralHumanFVV, our approach achieves the best performance in \textbf{PSNR},\textbf{SSIM},\textbf{LPIPS} and \textbf{MAE} metrics.}
 	\label{table:quantitative comparison}
 \end{table}
 
As we can see from the table, our HumanNeRF outperforms other methods across all metrics.
This demonstrates the generated views from our methods are closest to the real captured data. 
We also want to mention that even without per-scene fine-tuning, our method still achieves comparable results.

 \subsection{Ablation Study}
 
 \begin{figure*}[t]
	\begin{center}
		\includegraphics[width=1.0 \linewidth]{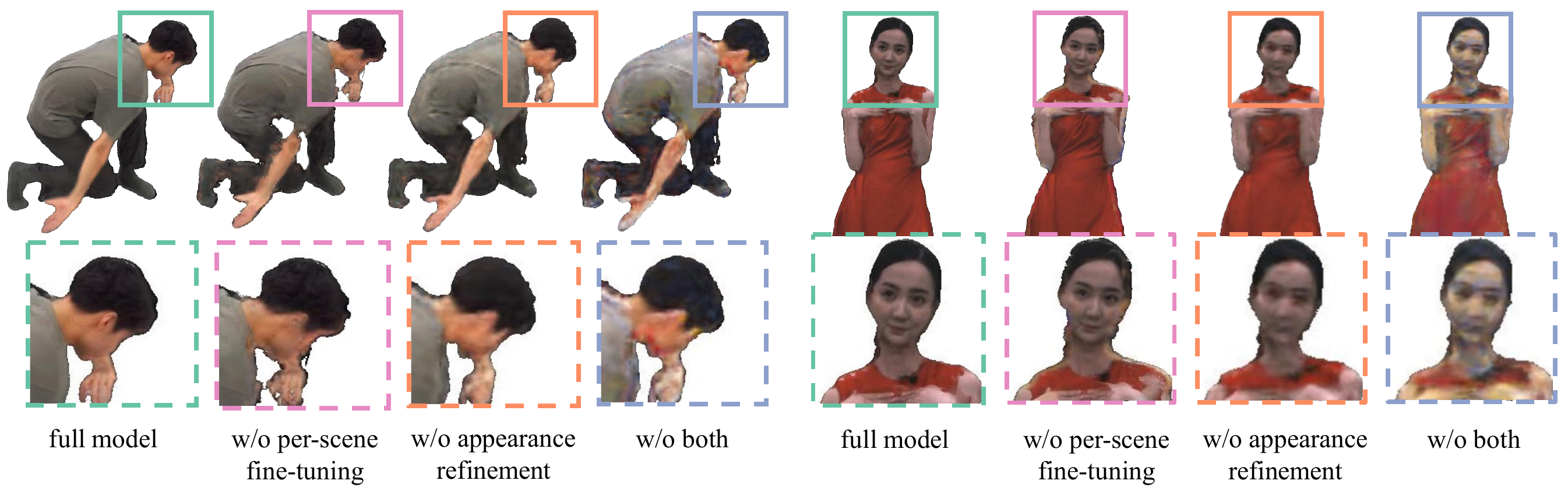}
	\end{center}
    \vspace{-0.4cm}
	\caption{Qualitative evaluation of different variations in our method. This evaluation demonstrates the contribution and effectiveness of our algorithmic components.}
	\label{fig:qualitative evaluation on three stage}
\end{figure*}

 \textbf{Appearance Blending and Fast Fine-tuning.} Here, we evaluate the performances of different modules in our approach. We first demonstrate the effectiveness of our per-scene fine-tuning strategy by directly comparing the output of generalizable NeRF and results after fine-tuning. As we can see in Fig.~\ref{fig:qualitative evaluation on three stage}, results without fine-tuning are low-detailed and blurry.  While lack of our novel appearance refinement module leads to blurring rendering artifacts, especially around the boundaries. In contrast, our complete approach achieves photo-realistic results with better decomposition for various entities. 
 
\textbf{Camera Number.} To evaluate the impact of the number of input views on our framework, we compare the results of our method with various numbers of input camera views. 

\begin{figure}[t]
	\begin{center}
		\includegraphics[width=0.95\linewidth]{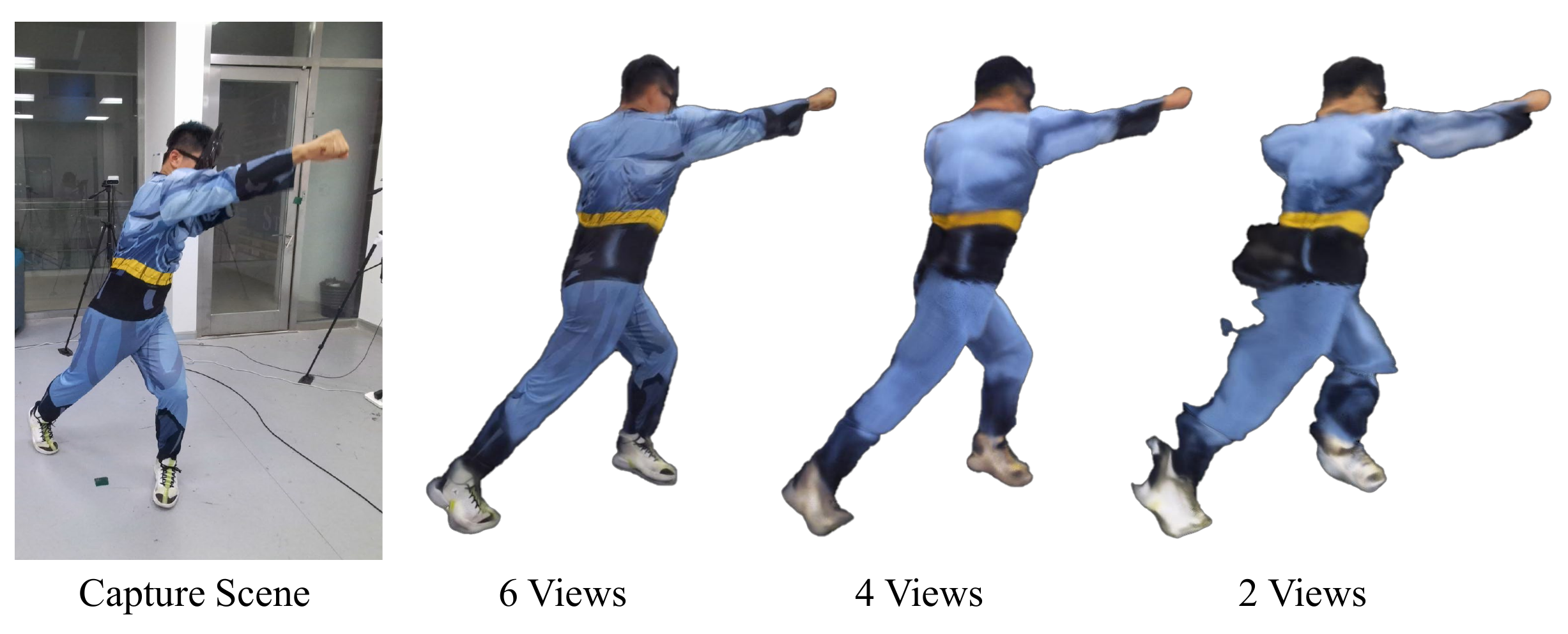}
	\end{center}
	\vspace{-0.4cm}
	\caption{ Evaluation of the number of input camera views. Our reconstructed appearance results using two, four and six cameras, respectively.}
	\label{fig:camera num evaluation}
\end{figure}

As shown in Fig.~\ref{fig:camera num evaluation}, the rendering results with views less than two suffer from severe geometric and rendering artifacts. We also use results generated with all cameras as the reference to calculate the corresponding PSNR, SSIM, and LPIPS. 

\begin{figure}[t]
	\begin{center}
		\includegraphics[width=1.0 \linewidth]{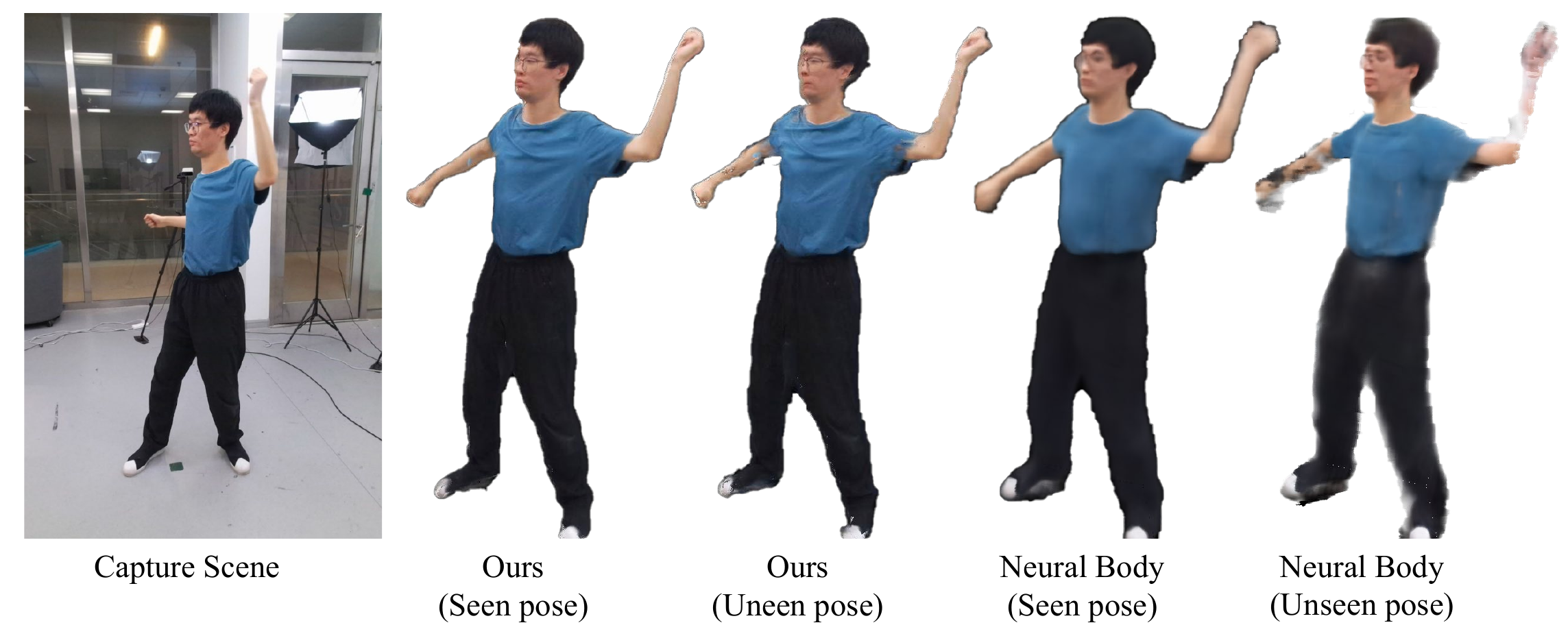}
	\end{center}
	\caption{\textbf{Qualitative evaluation on pose generalization.}}
	\label{fig:qualitative pose generalization}
\end{figure}

\textbf{Pose Generalization.} We further evaluate the pose generalizability of our HumanNeRF, we select 500 frames 'Swing' from videos in our multi-view dataset. We use 400 frames for fine-tuning and test on remaining 100 frames for pose generation. The results are shown in Fig.~\ref{fig:qualitative pose generalization}, our HumanNeRF generates visually good results even on the unseen pose and shows good metrics in Tab.~\ref{tb:quantitative pose generalization}.

\section{Conclusion}

We have presented a light weighted method for efficiently generating high-quality novel view synthesis of dynamic humans using only a sparse set of cameras. We leverage the fused image features and pose embedded human deformation module for dynamic human synthesize, and transcend the long-term per-scene optimization scheme of existing approaches. Moreover, our implicit neural appearance blending strategy refines results of volumetric rendering by borrowing fine details from two adjacent views. Experimental results on various datasets demonstrate the effectiveness of our approach in photo-realistic free-view synthesis even for challenging human poses and motions.	
With the ability of efficient generation, we believe that our approach may bring good insights to many critical applications in VR/AR, such as gaming, entertainment, education, immersive telepresence, etc.

\begin{table}[t]
	\centering
	\begin{tabular}{l|c|c|c|c}
		& PSNR$\uparrow$  & SSIM$\uparrow$ & LPIPS $\downarrow$ & MAE $\downarrow$ \\ \hline
		Ours(Seen) & \textbf{36.01} & \textbf{0.9897} & \textbf{0.0356} & \textbf{0.5963} \\
		Ours(Unseen) & 34.53  &0.9873  & 0.0386 & 0.7065\\
		NB(Seen) & 32.16  &0.9756  & 0.0626& 1.083\\
		NB(Unseen) & 27.61 & 0.9705 & 0.0640 & 1.756\\
		\bottomrule
	\end{tabular}
	\rule{0pt}{0.05pt}
	\caption{\textbf{Quantitative evaluation on pose generalization}.Results of ours and NB (Neural Body ) on the seen pose and unseen pose.}
	\label{tb:quantitative pose generalization}

\end{table}

\section{Acknowledgements}
This work was supported by Shanghai YangFan Program (21YF1429500), Shanghai Local college capacity building program (22010502800), NSFC programs (61976138, 61977047), the National Key Research and Development Program (2018YFB2100500), STCSM (2015F0203-000-06), and SHMEC (2019-01-07-00-01-E00003).

{\small
\bibliographystyle{ieee_fullname}
\bibliography{egbib}
}

\appendix
\newpage

\section{Limitations.}
In this paper, we propose a generalizable dynamic human neural radiance field method to
address issues of the existing approaches. Although very effective, the proposed HumanNeRF still needs hours of fine-tuning and has some limitations.
First, we use the regressed parametric human model to handle large pose deformation and complex motions, and it limits our approaches to the single-person setup and fails to handle the multi-person or human-object interaction situations.
Also though we have shown the generalization ability of our method, its capability is limited as distributions of human datasets only cover a small portion of the human dynamics and appearances.
Moreover, we do not explicitly model lighting conditions, significant brightness or color change between views may cause severe artifacts. For example, due to the switching of the nearest views during our appearance blending, jumping artifacts appear, especially for the significant brightness variance in our sparse input views. Such artifact will be alleviated if the illumination is almost consistent across views, as shown in the FVV results of the supplementary video.

\section{Components ablation study.} To better evaluate the components of our pipeline, we also do additional quantitative analysis of different modules of our method, such as without aggregated pixel alignment feature ($w/o\_F$), without pose embedded non-rigid human deformation ($w/o\_\mathbf{MLP}_d$),  and without neural appearance blending ($w/o\_ \mathbf{MLP}_\mathcal{A}$).
Note that our full module achieves the best results.
 \begin{table}[H]
 	\centering
 	\begin{tabular}{l|c|c|c|c}
 		 &  PSNR$\uparrow$ & SSIM$\uparrow$&LPIPS$\downarrow$ &MAE$\downarrow$ \\ \hline
 		Ours$_{wo\_F}$ & 18.36 &  0.8621 & 0.1503 & 13.49\\  
 		Ours$_{wo\_\mathbf{MLP}_d}$ & 26.79 &  0.9704 & 0.0516& 5.251\\
 		Ours$_{wo\_\mathbf{MLP}_\mathcal{A}}$ & 29.69 &  0.9620 & 0.0703& 2.016\\
 		Ours$_{full}$ & \textbf{33.01} &  \textbf{0.9842} & \textbf{0.0334} & \textbf{0.9307}\\    \hline
 	\end{tabular}
 	\rule{0pt}{0.05pt}
 	\caption{\textbf{Quantitative evaluation of different Components.}}
 	\label{table:Components}
 \end{table}
 
 As shown in Tab.~\ref{tb:qualitative camera num evaluation}, the average error increases rapidly as the camera number decreases. 
\begin{table}[H]
	\centering
	\begin{tabular}{l|c|c|c}
		& two views & four views & six views \\ \hline
		PSNR$\uparrow$ & 22.44 & 25.88 & \textbf{32.59}\\\
		SSIM$\uparrow$ & 0.9324  &0.9552  & \textbf{0.9817}\ \\
		LPIPS $\downarrow$ & 0.0887 & 0.0562 & \textbf{0.0304}\ \\
		\bottomrule
	\end{tabular}
	\rule{0pt}{0.05pt}
	
	\caption{\textbf{Quantity evaluation on the different number of input views}. We select six ,four and two camera views for ablation studies in \textbf{PSNR}, \textbf{SSIM} and \textbf{LPIPS} metrics.}
	\label{tb:qualitative camera num evaluation}
	
\end{table}
 
\section{Discussion about our generalizability .} 
\begin{figure}[H]
	\begin{center}
		\includegraphics[width=0.90\linewidth]{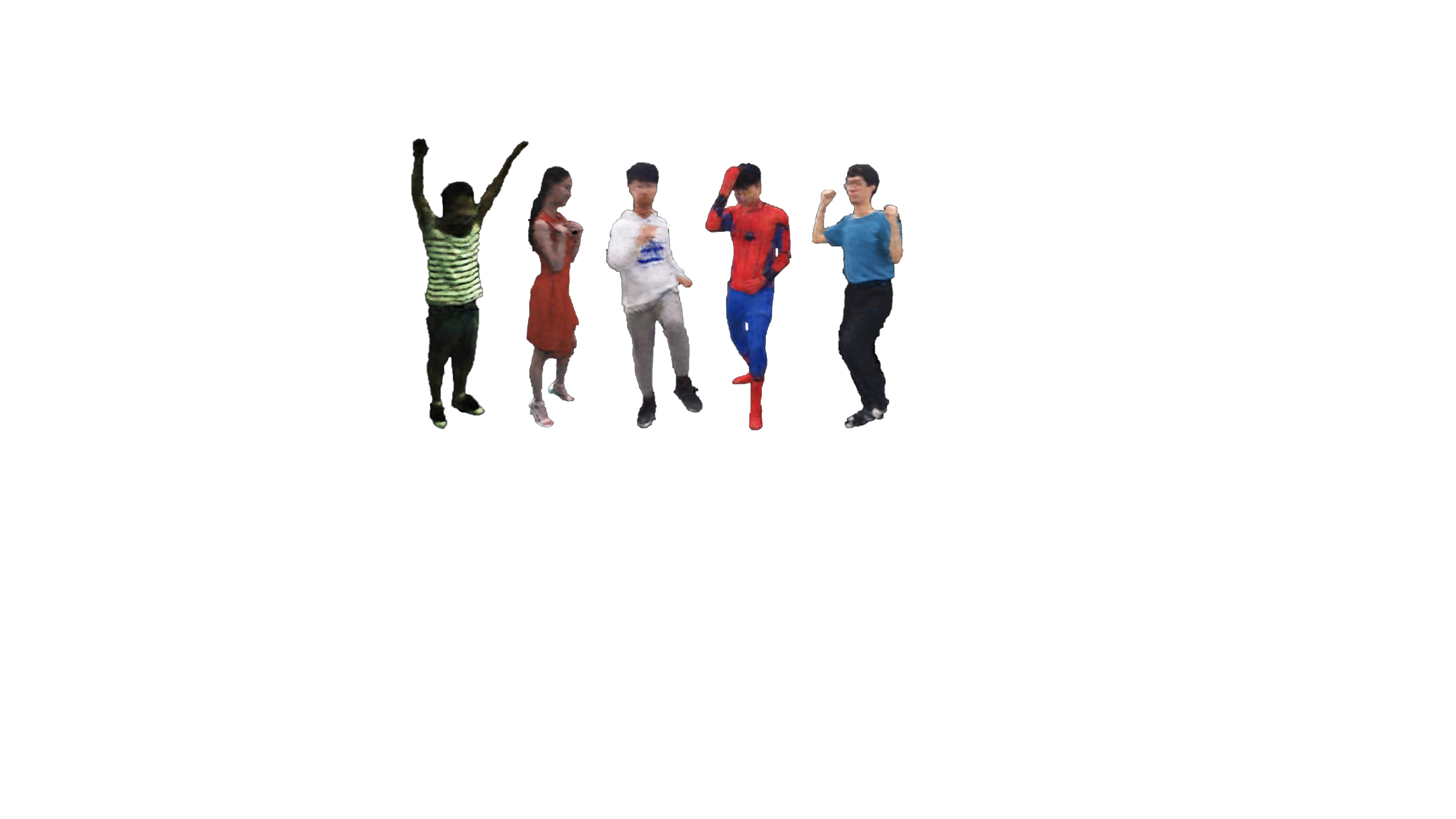}
	\end{center}
	\caption{Results of our generalizable dynamic neural radiance field module on unseen identities.}
	\label{fig:general}
\end{figure}

Despite the requirement of one hour fine-tuning of unseen identities, we would like to point out that our approach serves as a practical and more efficient scheme for dynamic and sparse view setting with significantly less fine-tuning effort than previous methods (see Tab.~\ref{table:time}). Our efficient generalizations are many-fold. First, only our generalizable NeRF module already provides meaningful yet blur results in Fig.~\ref{fig:general}, similar to Neural Human Performer [Kwon~\textit{et al.}].
Second, without per-scene fine-tuning, our method provides comparable results to previous general and even per-scene methods. Then, only with efficiently fine-tuning in hours, we can achieve SOTA performance, even for unseen poses.

 \section{Training time comparison.} 
 We compare our method with other per-scene traning methods in terms of training or fine-tuning time. As shown in Tab.~\ref{table:time}, our method is more efficient than other method.
  \begin{table}[H]
 	\centering
 	\begin{tabular}{l|c|c|c|c}
 		&  Ours & Neural Body & Neural Volumes & ST-NeRF\\ \hline
 		time & 1.2h & 6.7h & 8.4h& 9.5h\\ \hline
 	\end{tabular} 
 	\rule{0pt}{0.05pt}
 	\caption{Quantitative comparison against per-scene training methods in terms of \textbf{fine-tuning or training time} on the video "batman" with 300 frames of our multi-view dataset.}
 	\label{table:time}
 \end{table}
 
 \section{Network Architectures.} 
 We show detailed network architecture specifications of our feature extractor network $U$ (that extracts 2D image features ), feature blending network $\mathbf{MLP}_{\mathcal{B}}$, deformation network $\mathbf{MLP}_d$, generalizable dynamic neural radiance field $\Phi$ and appearance blending network $\mathbf{MLP}_\mathcal{A}$ .

\begin{table}[H]
\centering
\begin{tabular}{cccccc}
\toprule\hline
\textbf{Layer}  & \textbf{k} & \textbf{s} & \textbf{d}  & \textbf{channels} & input  \\ \hline
CRB2D Down$_0$ & 3 & 1 & 1 & $4/32$    & $I$\\
CRB2D Down$_1$ & 3 & 1 & 1 & $32/64$    & CRB2D Down$_0$ \\
CRB2D Down$_2$ & 3 & 1 & 1 & $64/128$   & CRB2D Down$_1$ \\
CRB2D Down$_3$ & 3 & 1 & 1 & $128/256$  & CRB2D Down$_2$ \\
CRB2D Up$_1$ & 3 & 1 & 1 & $256/256$  & CRB2D Down$_3$ \\
CRB2D Up$_2$ & 3 & 1 & 1 & $256/128$  & CRB2D Up$_1$ \\
CRB2D Up$_3$ & 3 & 1 & 1 & $128/64$  & CRB2D Up$_2$ \\
$T$       & 3 & 1 & 1 & $64/32$  & CRB2D Up$_3$ \\ \hline
\end{tabular}
\caption{Network details of feature extractor network $U$. \textbf{k} is the kernel size, \textbf{s} is the stride, \textbf{d} is the kernel dilation, and \textbf{channels} shows the number of input and output channels for each layer. We denote CRB2D to be ConvBnReLU2D.}
\vspace{-0.5cm}
\end{table}

\begin{table}[H]
\centering
\begin{tabular}{ccc}
\toprule\hline
\textbf{Layer}  & \textbf{channels} & input  \\ \hline
PE$_0$ &  $6/54,3/27$  & view direction $d,~$angle $\theta$\\

LR$_0$ &  $54+27+32*6/256$ & PE$_0,$ features $f$\\
LR$_1$ & $256/256$ & LR$_0$\\

LR$_2$ &  $256/256$  & LR$_1$\\
LR$_3$ &  $256/256$  & LR$_2$\\
LR$_4$ & $256/128$  & LR$_3$\\
LR$_5$ & $128/6$  & LR$_4$\\\hline
\end{tabular}
\caption{Network details of feature blending network $\mathbf{MLP}_{\mathcal{B}}$. PE/LR refers to the positional encoding and LinearRelu layer structure respectively (same as below).}
\vspace{-0.5cm}
\end{table}

\begin{table}[H]
\centering
\begin{tabular}{ccc}
\toprule\hline
\textbf{Layer}  & \textbf{channels} & input  \\ \hline
PE$_0$ & $24/216$  & $R_d$\\
LR$_0$ &  $216+72+32/256$ & PE$_0, R_v, F$\\
LR$_1$ & $256/256$ & LR$_0$\\

LR$_2$ &  $256/256$  & LR$_1$\\
LR$_3$ &  $256/256$  & LR$_2$\\
LR$_4$ & $256/128$  & LR$_3$\\
LR$_5$ & $128/3$  & LR$_4$\\\hline
\end{tabular}
\caption{Network details of deformation network $\mathbf{MLP}_d$. $R_d$ and  $R_v$ are the distances and directions between sample point $p$ and the 24 joints of the SMPL skeleton. $F$ is the feature after blending.}
\vspace{-0.5cm}
\end{table}

\begin{table}[H]
\centering
\begin{tabular}{ccc}
\toprule\hline
\textbf{Layer}   & \textbf{channels} & input  \\ \hline
PE$_0$ & 3/63 & position $x$ \\ 
LR$_0$ & 63/256 & PE$_0$ \\ 
LR$_1$ & 256/256 & LR$_0$\\ 
LR$_2$ & 256/256 & LR$_1$\\ 
LR$_3$ & 256/256 & LR$_2$\\ 
LR$_4$ & 27+256/256 & PE$_0$, LR$_3$\\ 
LR$_5$ & 256/256 & LR$_2$\\ 
LR$_6$ & 256/256 & LR$_2$\\ 
Density $\sigma$ & 256/1 & LR$_6$\\ 
PE$_1$ &  3/27 & view direction $d$\\ 
LR$_7$ & 256+27+32/256 & LR$_2, $PE$_1, F$\\ 
LR$_8$ & 256/128 & LR$_7$\\ 
Color $c$ & 128/3 & LR$_8$\\ \hline
\end{tabular}
\caption{Network details of generalizable dynamic neural radiance field $\Phi$.}

\end{table}

\begin{table}[H]
\centering
\begin{tabular}{ccc}
\toprule\hline
\textbf{Layer}   & \textbf{channels} & input  \\ \hline

LR$_0$     & $(32+3)*2 /256$ & $f_r, 0_r, f_l, 0_l$ \\ 
LR$_1$ &  $256/256$ & LR$_0$\\ 
LR$_2$ &  $256/256$  & LR$_1$\\
LR$_3$ &  $256/256$  & LR$_2$\\
LR$_4$ & $256/256$  & LR$_3$\\
LR$_5$ & $256/256$  & LR$_4$\\
LR$_6$ & $256/128$  & LR$_5$\\
Blending weights $W$ & $128/3$  & LR$_4$\\\hline

 \bottomrule
\end{tabular}
\caption{Network details of appearance blending network $\mathbf{MLP}_\mathcal{A}$. $f_r, 0_r, f_l, 0_l$ are two adjacent image features and occlusion maps.}

\end{table}

\end{document}